\documentclass[a4paper,conference]{IEEEtran}

\usepackage{cite}
\usepackage{graphicx}
\usepackage{xcolor}
\usepackage{booktabs}
\usepackage{amsmath,amsfonts,bm}
\usepackage{multirow}
\usepackage{amssymb}
\usepackage{hyperref}
\usepackage{cleveref}
\ifCLASSINFOpdf
\else
\fi

\begin{document}
\newcommand{\bi}[1]{\textcolor{red}{#1}}
\newcommand{\tb}[1]{\textbf{\textit{#1}}}

\title{Adaptive Soft Contrastive Learning}

\author{\IEEEauthorblockN{Chen Feng}
\IEEEauthorblockA{School of Electronic Engineering and Computer Science\\
Queen Mary University of London\\
London, UK\\
}
\and
\IEEEauthorblockN{Ioannis Patras}
\IEEEauthorblockA{School of Electronic Engineering and Computer Science\\
Queen Mary University of London\\
London, UK\\
}
}

% % make the title area
\maketitle
\IEEEpeerreviewmaketitle
\begin{abstract}
Self-supervised learning has recently achieved great success in representation learning without human annotations. 
The dominant method -- that is contrastive learning, is generally based on instance discrimination tasks, i.e., individual samples are treated as independent categories. However, presuming all the samples are different contradicts the natural grouping of similar samples in common visual datasets, e.g., multiple views of the same dog. To bridge the gap, this paper proposes an adaptive method that introduces soft inter-sample relations, namely Adaptive Soft Contrastive Learning (\tb{ASCL}). More specifically, ASCL transforms the original instance discrimination task into a multi-instance soft discrimination task, and adaptively introduces inter-sample relations.
As an effective and concise plug-in module for existing self-supervised learning frameworks, \tb{ASCL} achieves the best performance on several benchmarks in terms of both performance and efficiency. Code is available at \url{https://github.com/MrChenFeng/ASCL_ICPR2022}.

\end{abstract}


% \fi

% \tb{clustering: \cite{odc, deepcluster, sela, localAggregate, swav, pcl}}

% \tb{instance(explicitly):\cite{simclr, moco, instdisc}, implicitly:\cite{byol, simsiam, barlow, vicreg}}

% \tb{inter-sample: \cite{nnclr, bfnc, ifnd, meanshift, cld, dino, debiasedcl}}

% \tb{soft inter-relations: \cite{ressl, isd, clsa, sce, co2, sane}}

% \tb{ssl distill: \cite{seed, compress}}

\section{Introduction}
Self-supervised learning learns meaningful representation information through label-independent tasks, achieving performance that approaches or even exceeds that of supervised learning models in many tasks. Early self-supervised learning methods are often based on heuristic tasks, such as the prediction of image rotation angles, while the current mainstream methods are generally based on instance discrimination tasks, i.e., treating each individual instance as a separate semantic class. Methods in this category usually share the same framework, named as \tb{contrastive learning}. For a specific view of a specific instance, they define as positives other views of it and negatives views from other instances, and minimize its distance to positives while maximizing its distance to negatives. 
% With an feature encoder and a feature projector, along with a certain distance measure (cosine distance, L2-distance), and a certain number of negative samples, they expect to learn transformation invariant features by minimizing projection of views from same instance while maximizing projection of views from different instances.
Meanwhile, a large number of works have been done to improve this framework, such as using a momentum encoder and memory bank to increase the number of negatives~\cite{moco}. 

In this paper, we focus on an inherent deficiency of contrastive learning, namely ``class collision"~\cite{classcollsion,pcl}. The instance discrimination hypothesis violates the natural grouping in visual datasets and induces false negatives, e.g., the representation of two similar dogs should be close to each other rather than pushed away. To bridge the gap, we need to introduce meaningful inter-sample relations in contrastive learning. 
% Supervised contrastive learning~\cite{supcon} propose to utilize human annotations, however goes beyond fully self-supervised scenario. 

Debiased contrastive learning~\cite{debiasedcl} proposes a theoretical unbiased approximation of contrastive loss with the simplified hypothesis of the dataset distribution, however, does not address the issue of real false negatives. Some works~\cite{bfnc, ifnd} apply a progressive mechanism to identify and remove false negatives in the training. NNCLR~\cite{nnclr} tries to define extra positives for each specific view by ranking and extracting the top-$K$ neighbors in the learned feature space. Considering soft inter-sample relations, Co2~\cite{co2} introduces a consistency regularization enforcing relative distribution consistency of different positive views to all negatives. 
% As an extra regularization term for original contrastive loss, \cite{co2} is still affected by the inherent class collision problem. 
Clustering-based approaches~\cite{deepcluster, sela} also provide additional positives, but assuming the entire cluster is positive early in the training is problematic and clustering has an additional computational cost. 
In addition, all these methods rely on a manually set threshold or a predefined number of neighbors, which is often unknown or hard to determine in advance.

\begin{figure*}[htbp]
    \centering
    \includegraphics[width=0.95\textwidth]{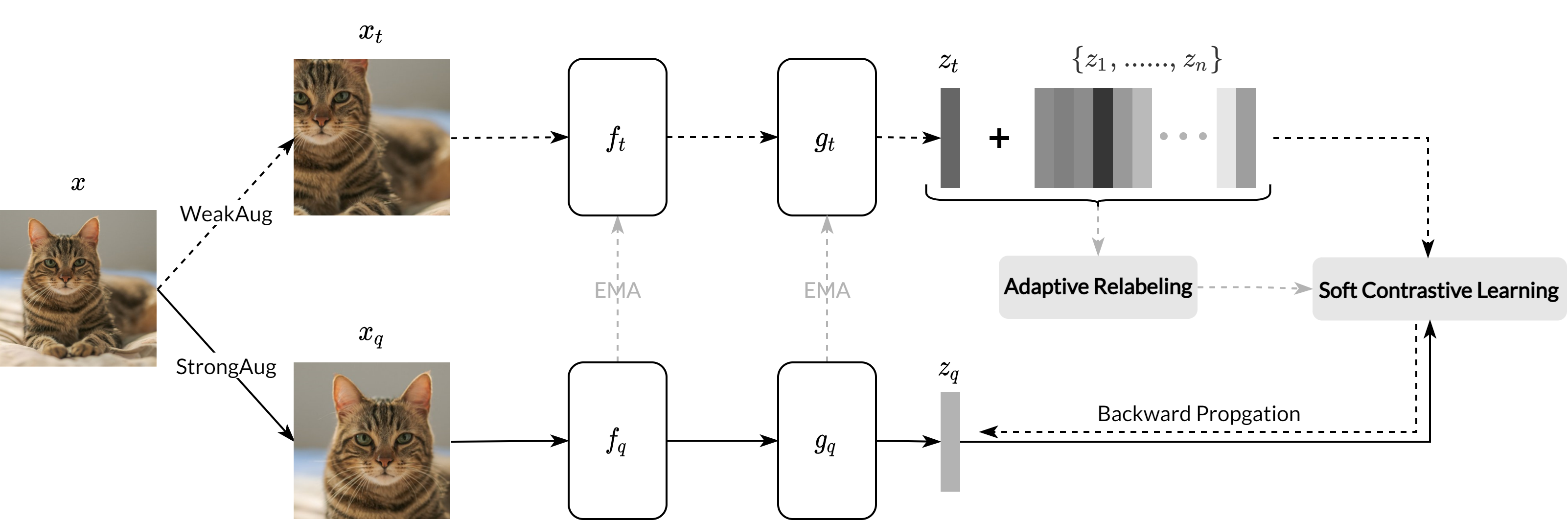}
    \caption{Structure of ASCL. When we remove the adpative relabelling step (indicated in light grey), ASCL can be considered as a general contrastive learning framework such as MoCo.}
    \label{fig:method}
\end{figure*}
In this work, we propose \tb{ASCL}, an efficient and effective module for current contrastive learning frameworks. We reformulate the contrastive learning problem and introduce inter-sample relations in an adaptive style. To make the training more stable and the inter-sample relationships more accurate, we use weakly augmented views to compute the relative similarity distribution and obtain the sharpened soft label information. 
Based on the uncertainty of the similarity distribution, we adaptively adjust the weights of the soft labels. In the early stages of training, due to the random initialization, the weights of the soft labels are low and the training of the model will be similar to the original contrastive learning.  As the features mature and the soft labels become more concentrated, the model will learn stronger inter-sample relations.

% To summarize, recent work has explored two main types of methods for introducing inter-sample information: hard considered more positive samples for each instance, which means converting the pseudo label as. 
% In supervised contrastive learning, the label is.
% in compress and SEED, the information comes from a pretrained encoder by knowledge distillation. However, we don't rely on a pretrained network, but directly distill knowledge from itself.

% In supervised learning tasks, knowledge distillation models have been widely shown to be effective by transferring learned inter-category relationships from teacher networks to student networks. Such inter-category relationships are also known as 'dark knowledge' in a supervised learning task. \cite{SEED,compress} take a similar approach by encoding the distance relationships between samples as distribution information via a softmax function and introducing a loss function based on KL-divergence. Without considering
% Meanwhile, label smoothing demonstrates that the student network has the same boosting effect on the teaching teacher network. 

The main contributions of this work are summarized as follows:
\begin{itemize}
    \item We propose a novel adaptive soft contrastive learning (\tb{ASCL}) method which smoothly alleviates the false negative and over-confidence in the instance discrimination task, and reduces the gap between instance-based learning with cluster-based learning;
    \item We show that weak augmentation strategies help to stabilize the contrastive training, both in our method and classical contrastive frameworks such as MoCo;
    \item We show that \tb{ASCL} keeps a high learning speed in the initial epochs compared to two other variants in our work that both try to introduce inter-sample relations in a hard style;
    \item Our method achieved the state of the art results in various benchmarks with very limited additional cost.
\end{itemize}

\section{Related Works}
\subsection{Early methods on self-supervised learning} 
Most of the early self-supervised learning methods rely on carefully designed heuristic tasks, such as jigsaw puzzles~\cite{jigsaw},  patch localization~\cite{patchloc}, image inpainting~\cite{colorization} and rotation prediction~\cite{rotationpred}. However, these pre-defined tasks lack enough relations to subsequent tasks such as image classification, and are outperformed by contrastive learning methods by a large margin currently.

\subsection{Contrastive learning with instance discrimination}
The idea of contrastive learning is first proposed in Hadsell et al.~\cite{dimensionreduction} and has become popular again as the dominant self-supervised learning method in recent years, achieving great performance close to or even exceeding that of fully supervised methods on many datasets. 
Exemplar network~\cite{exemplar} proposes to construct a $K$-way classifier for a dataset with $K$ images by treating every single image and its transformations as a unique surrogate class. Instance discrimination~\cite{instdisc} replaces the $K$-way classifier in~\cite{exemplar} with a non-parametric one consisting of all samples' representations and saved in a memory bank, to solve the problem of excessive memory requirement. 
% Since the memory bank in~\cite{instdisc} is not updated in real-time, \cite{invariant_siamise} apply siamese networks to accept both the original images and their corresponding augmentations as positive pairs and minimize the distance between them, while treating the input from different images as negative pairs. 
MoCo~\cite{moco} is an important baseline for current contrastive learning methods, which reuses the memory bank since samples in a single mini-batch may lead to insufficient negative pairs, and proposes a momentum encoder to update the memory bank in real-time to avoid outdated data representation. SimCLR~\cite{simclr} is another important baseline that finds that setting the mini-batch size to be large enough can eliminate the need for a memory bank. They also used stronger augmentation strategies and replaced the original single linear layer with an MLP.

Meanwhile, several works explore contrastive learning without negative samples. BYOL~\cite{byol} proposes an asymmetric network structure with a predictor and batch normalization to avoid mode collapse without explicit negative samples, while SimSiam~\cite{simsiam} shows that even a momentum encoder is not necessary. 
% With completely symmetric networks, Barlow twins~\cite{barlow} and VICReg\cite{vicreg} propose extra regularization terms on the variance of learned features, achieving results on par with the state of the art.

\subsection{Introducing inter-sample relations}
Most related to our work are recent works that explore how to introduce inter-sample relations into the original instance discrimination task. NNCLR~\cite{nnclr} builds on SimCLR by introducing a memory bank and searches for nearest neighbors to replace the original positive samples, while MeanShift~\cite{meanshift} relies on the same idea but builds on BYOL. Co2~\cite{co2} proposes an extra regularization term to ensure the relative consistency of both positive views with negative samples, while ReSSL~\cite{ressl} validates that the consistency regularization term itself is enough to learn meaningful representations.

\section{Adaptive Soft Contrastive Learning}
Current self-supervised learning methods focus on the instance discrimination task, more specifically, learning by considering each image instance as a separate semantic class.
In this work, we follow the representative structure in MoCo~\cite{moco}. More specifically, given a specific sample $x$, and two different transformed views of it, as query $x_q$ and target $x_t$, we want to minimize the distance of the corresponding representation projection $z_q$ and $z_t$ while maximizing the distance of $z_q$ and representations of other samples cached in a memory bank $\{z_1,..., z_n\}$. Here $z_{-}=g(f(x_{-}))$. The learned representation $f(x_{-})$ will be fixed and utilized in subsequent tasks such as image classification with an extra linear classifier~[Fig.~\ref{fig:method}]. With the encoders $f_q,f_t$ and projectors $g_q,g_t$, we optimize the infoNCE loss:
\begin{equation} \label{eq1}
    L = -log\frac{exp(z_q^Tz_t/\tau)}{exp(z_q^Tz_t/\tau)+\sum_{i=1}^n exp(z_q^Tz_i/\tau) }
\end{equation}

Where $\tau$ is a temperature hyperparameter that controls the feature density. 

\subsection{Soft contrastive learning}
Combining $z_t$ and memory bank $\{z_1,..., z_n\}$ together as $\{z'_1,z'_2,..., z'_{n+1}\} \triangleq \{z_t,z_1,..., z_n\}$\footnote{For the convenience, we may use these two notations interchangeably in the following.}, we can easily rewrite eq.~\ref{eq1} below:
\begin{equation} \label{eq2}
    L = -\sum_{j=1}^{n+1} y_jlogp_j
\end{equation}
where
\begin{equation} \label{eq3}
    p_j = \frac{exp(z_q^Tz'_j/\tau)}{\sum_{i=1}^{n+1} exp(z_q^Tz'_i/\tau)}
\end{equation}
\begin{equation} \label{eq4}
    y_j =\left\{
\begin{aligned}
& 1, ~j = 1\\
& 0, ~otherwise
\end{aligned}
\right.
\end{equation}
Here $\bm{y}=[y_1, ..., y_{n+1}]$ is the one-hot \tb{pseudo label} while $\bm{p}=[p_1, ..., p_{n+1}]$ is the corresponding prediction probability vector.  Recalling normal supervised learning, prediction over-confidence has inspired research on label smoothing and knowledge distillation. Similarly in self-supervised learning, this problem is more pronounced due to the fact that the distance between individual samples is smaller compared to that between categories, especially when there are duplicate samples or extremely similar samples in the dataset, i.e., the false negatives described earlier. By modifying \tb{pseudo label}, especially the part regarding with other samples, we can convert original contrastive learning problem as a soft contrastive learning problem, with the optimization goal in eq.~\ref{eq2}.

% In this work, we proposed \tb{ASCL} (Adaptive Soft Contrastive Learning) [Fig.~\ref{fig:method}]. By augmenting a simple adaptive soft relabelling mechanism to the current framework, \tb{ASCL} can effectively relieve the over-confidence and learn more meaningful inter-sample relations.

\subsection{Adaptive Relabelling}
As mentioned above, the \tb{pseudo label} in infoNCE loss ignores the inter-sample relations which will result in false negatives. To address this problem, we propose to modify the \tb{pseudo label} based on the neighboring relations in the feature space. We first calculate the cosine similarity $\bm{d}$ between self positive view $z'_1$ and other representations in memory bank $\{z'_2, z'_3, ..., z'_{n+1}\}$:
\begin{equation} \label{eq5}
    % d_i = \frac{\bm{z_t}^T \bm{z_i}}{ \| \bm{z_t}\|_{\small 2} \| \bm{z_i}\|_{\small 2}}
    d_j = \frac{{z'_1}^T z'_j}{ \| z'_1\|_{\small 2} \| z'_j\|_{\small 2}}, ~j=2,...,n+1
\end{equation}

\subsubsection{Hard relabelling}
According to $d_j, i=2,..., n+1$, we define the top-$K$ nearest neighbors set $\mathcal{N}_K$ in the memory bank of $z'_1$ as extra positives for $z_q$. The new \tb{pseudo label} $\bm{y}_{hard}$ will be defined as below: 
\begin{equation} \label{eq6}
    y_j =\left\{
\begin{aligned}
& 1, ~j = 1 \text{ or } z_j \in \mathcal{N}_K\\
& 0, ~otherwise
\end{aligned}
\right.
\end{equation}
Intuitively speaking, we consider not only $z'_1$ as positive for $z_q$ but also the top-$K$ nearest neighbors of $z'_1$.

\subsubsection{Adaptive hard relabelling}
However, it is risky to recklessly assume that the top-$K$ nearest neighbors are positive, and, especially early in the training, some hard samples may have fewer close neighbors compared to others. To alleviate these problems of $\bm{y}_{hard}$, we propose an adaptive mechanism that automatically modifies the confidence of the \tb{pseudo label}.
More specifically, with cosine similarity $\bm{d}$ we build the relative distribution $\bm{q}$ between self positive view $z'_1$ and other representations in memory bank $\{z'_2, z'_3, ..., z'_{n+1}\}$:
\begin{equation} \label{eq7}
    q_j = \frac{exp(d_j/\tau')}{\sum_{l=2}^{n+1} exp(d_l/\tau')},~j=2,...,n+1
\end{equation}
To quantify the uncertainty of relative distribution, i.e., how confident when we extract the neighbors, we define a confidence measure as the normalized entropy of the distribution $\bm{q}$:
\begin{equation} \label{eq8}
    c = 1 - \frac{H(\bm{q})}{log(n)}
\end{equation}
Here $H(\bm{q})$ is the Shannon entropy of $\bm{q}$. We further use $log(n)$ to normalize $c$ into $[0, 1]$.
We then get the adaptive hard label $\bm{y}_{ahcl}$ by augmenting $\bm{y}_{hard}$ with $c$:
\begin{equation} \label{eq9}
    y_j =\left\{
\begin{aligned}
& 1, ~j = 1 \\
& c, ~j\neq 1 \text{ and } z_j \in \mathcal{N}_K\\
& 0, ~j\neq 1 \text{ and } z_j \notin \mathcal{N}_K
\end{aligned}
\right.
\end{equation}

\subsubsection{Adaptive soft relabelling}
Moreover, instead of using top-$K$ neighbors for the extra positives, we also propose using the distribution $\bm{q}$ itself as soft labels. Intuitively speaking, a more concentrated distribution yields a higher degree of confidence, implying a more reliable neighboring relationship for the sample. We then define the adaptive soft label $\bm{y}_{ascl}$ as:
\begin{equation} \label{eq10}
    y_j =\left\{
\begin{aligned}
& 1, ~j = 1 \\
& min(1, ~c{\cdot}K{\cdot}q_j), ~j\neq 1 
\end{aligned}
\right.
\end{equation}
Here, $c$ is defined in eq.~\ref{eq8} to weight the soft labels, and $K$ is the number of neighbors in $\mathcal{N}_K$. Please note, that we put an upper bound of one -- that means that the most confident positive neighbor is not more confident than a view of the sample itself, i.e., than $z'_1$. Here, we 

Finally, $\bm{y}_{ascl}$, $\bm{y}_{ahcl}$ and $\bm{y}_{hard}$ are then normalized, that is:
% Please note, we also need the normalization operation for
\begin{equation} \label{eq11}
    y_j = \frac{y_j}{\sum_{-} y_{-}}
\end{equation}
For simplicity, we use the same notation for the normalized \tb{pseudo label} as the unnormalized ones. By default we use $\bm{y}_{ascl}$ for training --- this is the \tb{ASCL} method. We call the training method that uses $\bm{y}_{ahcl}$ as \tb{AHCL}, and the one with $\bm{y}_{hard}$ as \tb{Hard}. When we set $K$ as zero, the method degenerates to the original MoCo framework. 
% We first describe the adaptive relabelling mechanism of $y$. Motivated by self-distillation works, we 
% We adaptively relabel pseudo label $\mathbf{y}$:
% \begin{equation} \label{eq5}
%     y_k =\left\{
% \begin{aligned}
% & 1/(N+1), k = 1 \\
% & max(p_i, ), k = \{2,..., n+1\} \\
% \end{aligned}
% \right.
% \end{equation}
% where 
% certainty degree $\delta$ and relation prior $y'$. 

% \subsection{Soft contrastive learning}
% With certainty degree $\delta$ and relation prior $y'$, we follow the same . Please note, when we set $\delta = 1$ during the training, ASCL becomes normal MoCo [Fig.~\ref{fig:method}]. 

\subsection{Distribution sharpening}
The temperature $\tau$ in infoNCE loss~(\cref{eq1}) controls the density of the learned representations. Motivated by current semi-supervised learning works, to filter out possible noisy relations in the feature space we set a smaller temperature $\tau'$~(\cref{eq7}) for relative distribution $\bm{q}$ than $\tau$ in soft contrastive learning. We $\tau=0.1, \tau'=0.05$ by default. 

\subsection{Augmentation strategies}
Motivated by \cite{meanshift, ressl}, we also explore different augmentation strategies in \tb{ASCL}. Intuitively, strongly augmented samples have greater randomness and have larger errors in describing inter-sample relationships, while the use of weak augmentation leads to purer nearest-neighbor relationships, which in turn makes training more stable. In our approach, we use weak augmentation for the momentum encoder and memory bank, and strong augmentation for the online encoder. 

\subsection{ASCL without negative samples and memory bank}
%One may argue ASCL rely on necessary memory bank and explicit negative samples. As a flexible framework, we also apply \tb{ASCL} with BYOL~\cite{byol}, which is a representative self-supervised learning work of this kind. Following above notations of encoder and projector, with an extra predictor $h$, BYOL learn by enforcing the consistency between $z_t$ and $h(z_q)$. For more method details, please refer to Appendix.~\ref{appendixa}. 
In this section, we propose using ASCL in a contrastive learning framework that does not require a memory bank or use explicit negative samples. More specifically, so as to show that \tb{ASCL} is a flexible framework, we also apply \tb{ASCL} with BYOL~\cite{byol}, which is a representative self-supervised learning work. BYOL uses an extra predictor $h$ and learns by enforcing the consistency between $h(z_q)$ and $z_t$. We extend this by considering all samples in the batch, labeling them softly according to the distance from the sample in question, and optimizing the consistency according to the soft labels. For more details, please refer to Appendix~\ref{appendixa}.

\begin{table*}[htbp]
\centering
\caption{Results on small-scale and medium-scale datasets.}
% \resizebox{\columnwidth}{!}{%
\begin{tabular}{lcccccc}
\hline
Method                 & BackProp & EMA & CIFAR-10       & CIFAR-100      & STL-10         & Tiny ImageNet  \\ \hline
Supervised             & -        & No  & 94.22          & 74.66          & 82.55          & 59.26          \\ \hline
SimCLR~\cite{simclr}   & 2x       & No  & 84.92          & 59.28          & 85.48          & 44.38          \\
BYOL~\cite{byol}       & 2x       & Yes & 85.82          & 57.75          & 87.45          & 42.70          \\
SimSiam~\cite{simsiam} & 2x       & No  & 88.51          & 60.00          & 87.47          & 37.04          \\
MoCo~\cite{moco}       & 1x       & Yes & 86.18          & 59.51          & 85.88          & 43.36          \\
ReSSL~\cite{ressl}     & 1x       & Yes & 90.20          & 63.79          & 88.25          & 46.60          \\
ReSSL(*)               & 1x       & Yes & 90.23          & 64.31          & 87.69          & 45.94          \\
ASCL(Ours)             & 1x       & Yes & \textbf{90.55} & \textbf{65.27} & \textbf{89.54} & \textbf{48.36} \\ \hline
\end{tabular}
% }
\label{tab:small}
\end{table*}

% % \subsection{Relation to other works}
% % \tb{MoCo} sss

% % \paragraph{Relabelling based on clustering}
% % SwAV~\cite{swav}, Deepcluster~\cite{deepcluster}, SeLa~\cite{sela} applied the clustering idea for. More specifically,  

% % \paragraph{Self-supervised knowledge distillation}

\section{Experiments and results}
\subsection{Experiment settings}

\subsubsection{Datasets} 
\paragraph{CIFRA10 and CIFAR100} Both CIFAR10 and CIFAR100 consist of 50K training images and 10K test images with 32$\times$32 pixel image resolution. CIFAR10 has 10 classes while CIFAR100 has 100 classes.
\paragraph{STL10} STL10 consists of 100K unlabeled images, 5K labeled training images, and 8K test images with 96$\times$96 image resolution and 10 classes.
\paragraph{Tiny ImageNet} Tiny ImageNet consists of 100K training images and 10K validation images with 200 classes. Tiny ImageNet is a lite version of ImageNet with image resizing to 64x64 pixels.
\paragraph{ImageNet-1k} ImageNet-1K is a large dataset with almost 1.3M images in the training set and 50K images in the validation set, also known as the ILSVRC-2012 dataset.

\subsubsection{Implementation details}
For small-scale datasets: CIFAR10, CIFAR100, STL10, and Tiny ImageNet, we apply ResNet-18 as the backbone. To adapt to the low image resolution, we modify the ResNet-18 structure by modifying the first convolutional layer and removing the max-pooling layer. For ImageNet-1k, we apply ResNet-50 as backbone. 

For small-scale datasets we set the memory bank size as 4096 while for ImageNet-1k as 65536. The updating momentum for memory encoder is 0.99. For data augmentations, we apply strong augmentations consisting of a random resized crop~(range from 0.2 to 1.0, for CIFAR10 and CIFAR100, size 32$\times$32, for STL10 and Tiny ImageNet, size 64$\times$64, for ImageNet-1K, size 224$\times$224), horizontal flip~(with probability=0.5), color distortion~(with strength=0.8), Gaussian blur~(with probablity=0.5) and grayscale~(with probablity=0.2). For weak augmentations we only keep random resized crop and horizontal flip. For model hyperparameters, we set $K=1$, $\tau=0.1$ and $\tau'=0.05$ by default. 

% For small-scale datasets, the representation dimension is 128 while the projector is a 3-layer MLP with a 512-dimension output. For ImageNet-1k, the representation dimension is 
We train the network with SGD optimizer for 200 epochs with a momentum of 0.9 and weight decay of 1e-4. The initial learning rate is 0.06 and is controlled by a cosine annealing scheduler. The batchsize is fixed as 256. Optional batchsize should be equipped with learning rate = $0.06\times batchsize / 256$.

\subsubsection{Evaluation protocol}
We freeze the parameters of encoder $f_{-}$ and train a linear classifier without the projector $g_{-}$. We train the classifier with SGD optimizer for 100 epochs with a momentum of 0.9 and no weight decay. The initial learning rate is 10 and reduced to 1 and 0.1 at the 60th and 80th epochs, respectively. Especially for STL10, we pretrain with both 100K unlabeled data and 5K labeled training images, while for evaluation, we train with only the 5K labeled training images and test on the 8K test images. For online KNN evaluation, we extract the representation for all train samples and apply a distance-weighted KNN classification on the test samples.

\subsection{Results on small-scale and medium-scale datasets}
We evaluate our method on small-scale datasets and compare with baselines, in Table ~\ref{tab:small}. The results of other methods are copied from the recent work~\cite{ressl} with their best results. For fair comparison, we also report the reproduced results of \cite{ressl}, noted with star marker. It is clear that we perform the best compared to the baselines.

% \begin{table}[htbp]
% \centering
% \caption{Results on small-scale and medium-scale datasets.}
% \resizebox{\columnwidth}{!}{%
% \begin{tabular}{@{}lccccc@{}}
% \toprule
% Method     & BackProp        & CIFAR-10       & CIFAR-100      & STL-10         & Tiny ImageNet  \\ \midrule
% Supervised & -                 & 94.22          & 74.66          & 82.55          & 59.26          \\ \midrule
% SimCLR~\cite{simclr}     & 2x               & 84.92          & 59.28          & 85.48          & 44.38          \\
% BYOL~\cite{byol}       & 2x               & 85.82          & 57.75          & 87.45          & 42.70          \\
% SimSiam~\cite{simsiam}    & 2x               & 88.51          & 60.00          & 87.47          & 37.04          \\
% MoCo~\cite{moco}       & 1x              & 86.18          & 59.51          & 85.88          & 43.36          \\
% ReSSL~\cite{ressl}      & 1x               & 90.20          & 63.79          & 88.25          & 46.60          \\
% ReSSL(*)   & 1x              & 90.23          & 64.31          & 87.69          & 45.94          \\
% ASCL(Ours) & 1x              & \textbf{90.55} & \textbf{65.27} & \textbf{89.54} & \textbf{48.36} \\ \bottomrule
% \end{tabular}
% }
% \label{tab:small}
% \end{table}

\subsection{Ablations study}

\begin{figure*}[htbp]
    \centering
    \includegraphics[width=0.7\textwidth]{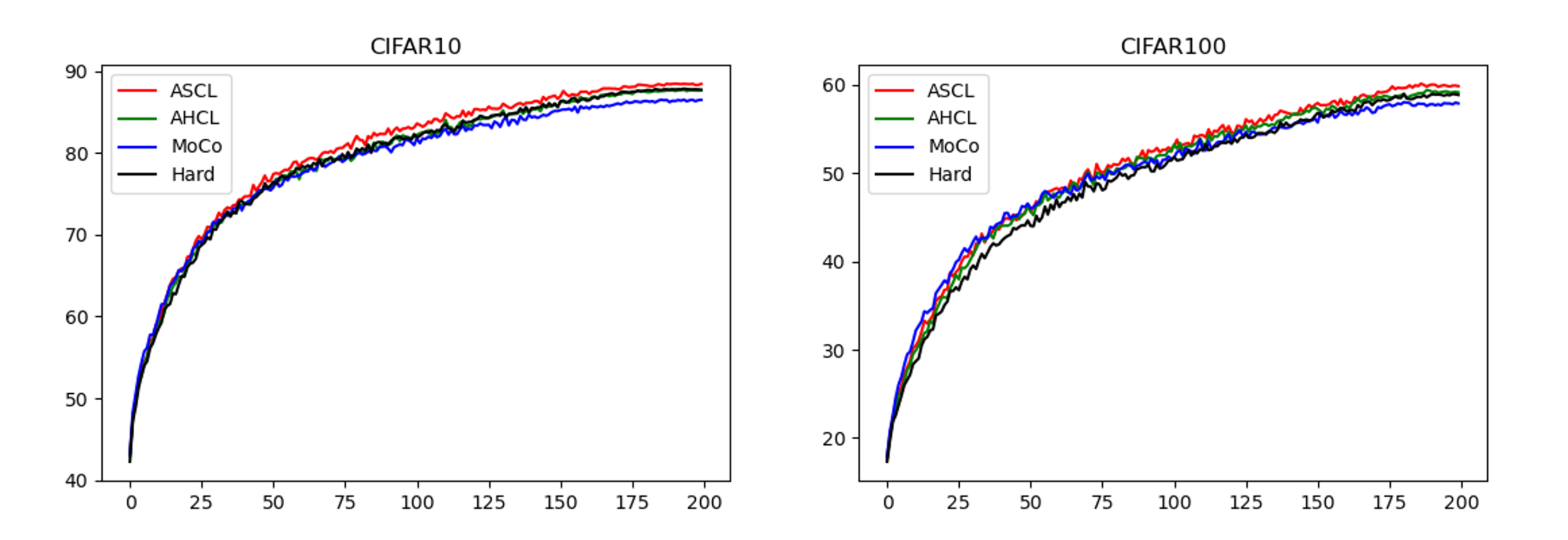}
    \caption{KNN online evaluation accuracy.}
    \label{fig:knnacc}
\end{figure*}

\begin{figure*}[htbp]
    \centering
    \includegraphics[width=0.7\textwidth]{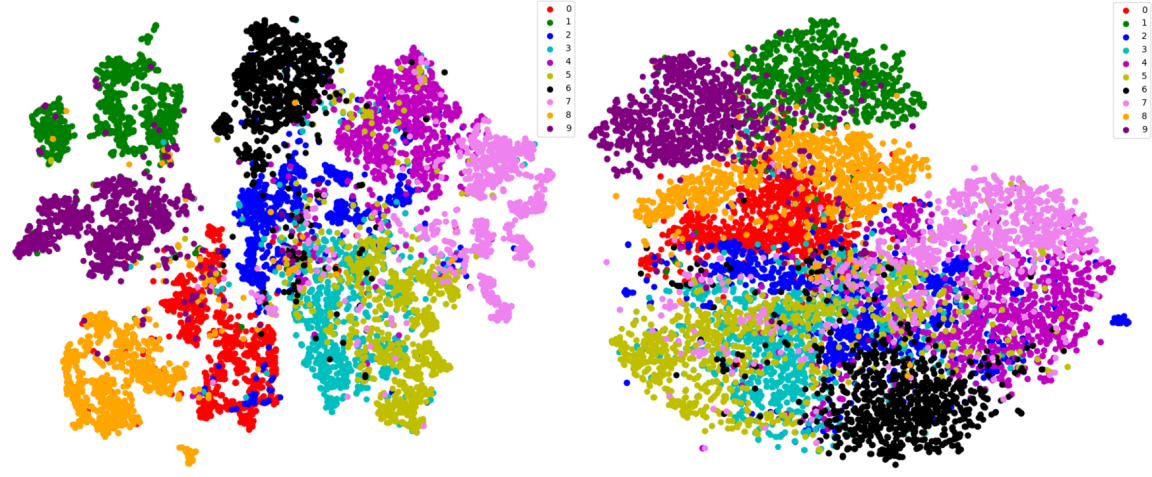}
    \caption{t-SNE visualization of learned features on CIFAR10, classes indicated by different colors. Left: ASCL, Right: MoCo. }
    \label{fig:tsne}
\end{figure*}
\subsubsection{Effect of augmentation strategies}
Strong augmentation strategies help model to learn transformation-invariant representations. Some works explore even stronger augmentations for further improvement~\cite{clsa}. By contrast, in \tb{ASCL} we apply a weak augmented view for momentum encoder and memory bank to stabilize the inter-sample relations. We find a stable memory bank with weak augmentations benefits learning.  Since the lower performance may come from the slower convergence induced by strong augmentation, we train the model for more epochs to further validate the effect of the weak augmentations.
\begin{table}[htbp]
\centering
\caption{Effect of weak augmentations.}
\label{tab:augmentation}
\resizebox{\columnwidth}{!}{%
% \begin{tabular}{@{}lcccc@{}}
% \toprule
% \multirow{2}{*}{Method} & \multicolumn{2}{c}{CIFAR10} & \multicolumn{2}{c}{CIFAR100} \\ \cmidrule(l){2-5} 
%                         & 200 epochs   & 800 epochs   & 200 epochs    & 800 epochs   \\ \midrule
% MoCo                    & 88.55         &   91.98           & 64.02                  &              \\
% ASCL                    & \textbf{90.55}        &              & \textbf{65.46}         &              \\ \bottomrule
% \end{tabular}}
\begin{tabular}{@{}cccccc@{}}
\toprule
\multicolumn{2}{c}{\multirow{2}{*}{Method}} & \multicolumn{2}{c}{CIFAR10} & \multicolumn{2}{c}{CIFAR100} \\ \cmidrule(l){3-6} 
\multicolumn{2}{c}{}  & 200 epochs   & 800 epochs   & 200 epochs    & 800 epochs   \\ \midrule
\multirow{2}{*}{MoCo} & Strong & 83.48 & 89.61 & 57.98 &  64.00 \\
                      & Weak   & 88.55 & 91.98 & 64.02 &  68.16 \\ \midrule
\multirow{2}{*}{ASCL} & Strong & 85.97 & 91.45 & 59.34 &  66.85 \\
                      & Weak   & \textbf{90.00} & \textbf{92.84} & \textbf{65.46} & \textbf{69.19} \\ \bottomrule
\end{tabular}}
\end{table}

In Table~\ref{tab:augmentation}, we can see that for both \tb{ASCL} and original MoCo, a more stable memory bank is always beneficial to bring consistent improvement, while \tb{ASCL} always performs better than MoCo in all settings. 

\subsubsection{Learning speed analysis}
In Fig.~\ref{fig:knnacc} we show the online KNN evaluation accuracy of the original MoCo and three variants of our method. As expected, all three variants surpass the original MoCo on both CIFAR10 and CIFAR100 datasets considering the KNN accuracy. However, previous works~\cite{nnclr, meanshift} usually experience a slower learning speed as they described that introducing neighbors makes the task harder. This is validated by the KNN accuracy trend of the \tb{Hard} variant of our method. However, both \tb{ASCL} and \tb{AHCL} keep a high learning speed similar to the original MoCo which implies the adaptive mechanism makes the framework more reliable and efficient, bringing risk-free improvement even when trained with few epochs.

\subsubsection{Representation visualizations}
In Fig.~\ref{fig:tsne}, we compare the representations learned by our method with those learned by MoCo. Different classes of samples are mixed with each other in the learned presentation space of MoCo while we can find clear boundaries in the \tb{ASCL} representations. It is clear that the relative relationship of different classes is significantly improved with \tb{ASCL}. 
% \begin{figure}[htbp]

\begin{table*}[htbp]
\centering
\caption{Effect of different number of neighbors $K$.}
\label{tab:nns}
% \resizebox{\columnwidth}{!}{%
% \begin{tabular}{@{}lcccccccc@{}}
% \toprule
% \multirow{2}{*}{Method} & \multicolumn{4}{c}{CIFAR10}   & \multicolumn{4}{c}{CIFAR100}  \\ \cmidrule(l){2-9} 
%                          & $K=1$   & $K=2$   & $K=5$   & $K=10$  & $K=1$   & $K=2$   & $K=5$   & $K=10$  \\ \midrule
% % MoCo  & \multicolumn{4}{c}{83.48}     & \multicolumn{4}{c}{57.98}     \\
% MoCo                     & \multicolumn{4}{c}{88.55}     & \multicolumn{4}{c}{64.02}     \\ \midrule
% Hard                     & 89.52 & 89.99 & 90.06 & 89.99 & 64.19 & 64.08 & 63.82 & 63.71 \\
% AHCL                     & 89.60 & 89.85 & 89.92 & 89.89 & 64.98 & 64.72 & 64.43 & 63.94 \\
% ASCL                     & \textbf{90.00} & \textbf{90.09} & \textbf{90.55} & \textbf{90.27} & \textbf{65.46} & \textbf{65.35} & \textbf{64.83} & \textbf{64.39} \\ \bottomrule
% \end{tabular}
\begin{tabular}{@{}lcccccccc@{}}
\toprule
\multirow{2}{*}{Method} & \multicolumn{4}{c}{CIFAR10}   & \multicolumn{4}{c}{CIFAR100}  \\ \cmidrule(l){2-9} 
                         & $K=1$   & $K=2$   & $K=5$   & $K=10$  & $K=1$   & $K=2$   & $K=5$   & $K=10$  \\ \midrule
% MoCo  & \multicolumn{4}{c}{83.48}     & \multicolumn{4}{c}{57.98}     \\ \midrule
MoCo                     & \multicolumn{4}{c}{88.55}     & \multicolumn{4}{c}{64.02}     \\ \midrule
Hard                     & 89.52 & 89.99 & 90.06 & 89.99 & 64.19 & 64.08 & 63.82 & 63.71 \\
AHCL                     & 89.60 & 89.85 & 89.92 & 89.89 & 64.98 & 64.72 & 64.43 & 63.94 \\
ASCL                     & 90.00 & 90.09 & \textbf{90.55} & 90.27 & \textbf{65.46} & 65.35 & 64.83 & 64.39 \\ \bottomrule
\end{tabular}
% }
\end{table*}

\begin{table*}[htbp]
\caption{}
\centering
\label{tab:imagenet}
\begin{tabular}{lcccccc}
\hline
Method                         & Architecture & BackProp & EMA & Batch Size & Epochs & Top-1 Acc     \\ \hline
Supervised                     & ResNet50     & 1x       & No  & 256        & 120    & 76.5          \\ \cline{2-7} 
InstDisc~\cite{instdisc}       & ResNet50     & 1x       & No  & 256        & 200    & 58.5          \\
LocalAgg~\cite{localAggregate} & ResNet50     & 1x       & NO  & 128        & 200    & 58.8          \\
MoCo~\cite{moco}               & ResNet50     & 1x       & Yes & 256        & 200    & 67.5          \\
CO2~\cite{co2}                 & ResNet50     & 1x       & No  & 256        & 200    & 68.0          \\
PCL~\cite{pcl}                 & ResNet50     & 1x       & Yes & 256        & 200    & 67.6          \\
ReSSL~\cite{ressl}             & ResNet50     & 1x       & Yes & 256        & 200    & 69.9          \\
ASCL(Ours)                     & ResNet50     & 1x       & Yes & 256        & 200    & \textbf{71.5} \\ \hline
SimCLR~\cite{simclr}           & ResNet50     & 2x       & No  & 4096       & 200    & 66.8          \\
NNCLR~\cite{nnclr}             & ResNet50     & 2x       & No  & 4096       & 200    & 70.7          \\
CLSA~\cite{clsa}               & ResNet50     & 2x       & Yes & 256        & 200    & 69.4          \\
SwAV~\cite{swav}               & ResNet50     & 2x       & No  & 4096       & 200    & 69.1          \\
SimSiam~\cite{simsiam}         & ResNet50     & 2x       & No  & 256        & 200    & 70.0          \\
BYOL~\cite{byol}               & ResNet50     & 2x       & Yes & 4096       & 200    & 70.6          \\ \hline
\end{tabular}
% \begin{tabular}{@{}lcccc@{}}
% \toprule
% Method     & BackProp & Batch Size & Epochs & Top-1 Acc     \\ \midrule
% Supervised & 1x       & 256        & 120    & 76.5          \\ \midrule
% InstDisc~\cite{instdisc}   & 1x       & 256        & 200    & 58.5          \\
% LocalAgg~\cite{localAggregate}   & 1x       & 128        & 200    & 58.8          \\
% MoCo~\cite{moco}       & 1x       & 256        & 200    & 67.5          \\
% CO2~\cite{co2}        & 1x       & 256        & 200    & 68.0          \\
% PCL~\cite{pcl}        & 1x       & 256        & 200    & 67.6          \\
% ReSSL~\cite{ressl}      & 1x       & 256        & 200    & 69.9          \\
% ASCL(Ours) & 1x       & 256        & 200    & \textbf{71.5} \\ \midrule
% SimCLR~\cite{simclr}     & 2x       & 4096       & 200    & 66.8          \\
% NNCLR~\cite{nnclr}      & 2x       & 4096       & 200    & 70.7          \\
% CLSA~\cite{clsa}       & 2x       & 256        & 200    & 69.4          \\
% SwAV~\cite{swav}       & 2x       & 4096       & 200    & 69.1          \\
% SimSiam~\cite{simsiam}    & 2x       & 256        & 200    & 70.0          \\
% BYOL~\cite{byol}       & 2x       & 4096       & 200    & 70.6          \\ \bottomrule
% \end{tabular}
\end{table*}

\begin{table}[htbp]
\centering
\caption{Effect of different temperature $\tau'$.}
\label{tab:tempscan}
\resizebox{\columnwidth}{!}{%
\begin{tabular}{@{}lccccc@{}}
\toprule
Dataset  & $\tau' = 0.01$ & $\tau' = 0.02$ & $\tau' = 0.05$ & $\tau' = 0.08$ & $\tau' = 0.1$ \\ \midrule
CIFAR10  &    89.67          &   89.82           &      \textbf{90.00}      &      88.91        &    88.58         \\
CIFAR100 &    64.94          &   64.69        &      \textbf{65.46}    &    64.26  &        64.32     \\ \bottomrule
\end{tabular}
}
\end{table}

\subsubsection{Appropriate sharpening of inter-sample relations is beneficial}
In most contrastive learning algorithms, the temperature parameter $\tau$ is very critical. In \tb{ASCL}, we use an extra $\tau'$ to sharpen the relative distribution~(eq.~\ref{eq7}), thus to remove the possible noisy inter-sample relations but focus on the most important one. In Table~\ref{tab:tempscan}, we fix $\tau= 0.1$ and perform extensive experiments on CIFAR10 and CIFAR100 with $\tau' = \{0.01, 0.02, 0.05, 0.08, 0.1\}$. It is clear that a too low or too high $\tau'$ is not optimal, but still better than the original MoCo (88.55\% for CIFAR10, 64.02\% for CIFAR100), which again shows the robustness of \tb{ASCL}. Intuitively speaking, when $\tau'\rightarrow 0$, \tb{ASCL} is equivalent to the \tb{Hard} with $K = 1$, and when $\tau'$ increases, our method is equivalent to introducing uniform distribution, similar to label smoothing in supervised learning.

\subsubsection{Robustness to number of neighbors}
In Table~\ref{tab:nns} we evaluate our methods with different $K$ --- number of neighbors. Please note, for a fair comparison, we report MoCo with weak augmentations for the momentum encoder here. We can find that for \tb{Hard} and \tb{AHCL}, too many neighbors ($K=10$ for example) result in reduced performance compared to MoCo (which equals introducing no neighbors). The adaptive mechanism helps as \tb{AHCL} is always better than \tb{Hard} while \tb{ASCL} is more robust achieving the best results in all conditions, with further soft pseudo labels.

\subsection{Results on ImageNet-1k}
We also evaluate \tb{ASCL} on ImageNet-1k in Table~\ref{tab:imagenet}. With all methods pretrained for 200 epochs, \tb{ASCL} outperforms the current state-of-the-art methods. Also, please note that \tb{ASCL} requires only one backpropagation pass, which reduces a significant amount of computational cost compared to methods such as BYOL, SimCLR, etc.

\subsection{ASCL without negative pairs and memory bank}
In Table~\ref{tab:byol}, we compare the performance of BYOL and BYOL with \tb{ASCL}~(\cref{appendixa}).
% On CIFAR10 ASCL brings significant improvements, however, for CIFAR100 dataset, ASCL does hard to original BYOL performance, which. 
\tb{ASCL} gets better and worse performance on CIFAR10 and CIFAR100, respectively. Intuitively, we conjecture that the risk of introducing false positives outweighs the benefits of extra positives. Specifically, for CIFAR100 dataset, we set the batchsize to 256 which is relatively small considering there are 100 different semantic classes. To verify this hypothesis, we increase the batch size to 512 and find that the gap between the two was significantly reduced.

\begin{table}[htbp]
\centering
\caption{Performance of ASCL with BYOL.}
\label{tab:byol}
\resizebox{\columnwidth}{!}{%
\begin{tabular}{@{}lccc@{}}
\toprule
Method &
  \begin{tabular}[c]{@{}c@{}}CIFAR10 \\ (batchsize = 256)\end{tabular} &
  \begin{tabular}[c]{@{}c@{}}CIFAR100 \\ (batchsize = 256)\end{tabular} &
  \begin{tabular}[c]{@{}c@{}}CIFAR100 \\ (batchsize = 512)\end{tabular} \\ \midrule
BYOL~\cite{byol} &
  86.95 &
  \textbf{62.31} &
  \textbf{61.03} \\
BYOL + ASCL &
  \textbf{90.49} &
  58.51 &
  60.37 \\ \bottomrule
\end{tabular}
}
\end{table}

\section{Conclusions}
In this work, we propose \tb{ASCL}, a reliable and efficient framework based on the current contrastive learning framework. We utilize a sharpened inter-sample distribution to introduce extra positives and adaptively adjust its confidence based on the entropy of the distribution. Our method achieves the state of the art in various benchmarks, with a negligible extra computational cost. We also show the potential of our method with self-supervised learning methods requiring no memory bank and explicit negative pairs.

{\setlength{\parindent}{0cm}
\textbf{Acknowledgments:} This work was supported by the EU H2020 AI4Media No. 951911 project.
}

\bibliographystyle{IEEEtran}

% \onecolumn
\appendix

\subsection{Details of ASCL with BYOL}
\label{appendixa}
We here explain how to apply \tb{ASCL} with BYOL in detail. For implementation details of the BYOL, please refer to the original paper~\cite{byol}.
With a mini-batch of samples as ${x_1, x_2,..., x_b}$ and respective feature projections ${z^1_t, z^2_t,..., z^b_t}$ (from momentum encoder) and ${z^1_q, z^2_q,..., z^b_q}$ (from online encoder), BYOL optimizes the following loss for each sample $x_i$:
\begin{equation}
    L = 2 - 2 \cdot \frac{{z^i_t}^T h(z^i_q)}{ \| z^i_t\|_{\small 2} \| h(z^i_q)\|_{\small 2}}
\end{equation}
We can easily reformulate it as below:
\begin{equation}
    L = 2 - 2 \cdot \sum_{j=1}^b y_j \frac{{z^i_t}^T h(z^j_q)}{ \| z^i_t\|_{\small 2} \| h(z^j_q)\|_{\small 2}}
\end{equation}
With the \tb{pseudo label} $\bm{y}$ denoting inter-sample relations as:
\begin{equation} \label{eq12}
    y_j =\left\{
\begin{aligned}
& 1, ~j = i \\
& 0, ~j=\{1,..., b\} \setminus i \\
\end{aligned}
\right.
\end{equation}
We then apply \tb{ASCL} to build inter-sample relations directly based on the samples of each mini-batch. For sample $x_i$, we calculate the cosine similarity between $z^i_t$ and all other projections $\{z^1_t, z^2_t,..., z^b_t\}\setminus z^i_t$ in the mini-batch. 
% For consistency, we abuse the notations with 
% $z^i_t \triangleq z_t$, while 
\begin{equation} \label{eq13}
    % d_i = \frac{\bm{z_t}^T \bm{z_i}}{ \| \bm{z_t}\|_{\small 2} \| \bm{z_i}\|_{\small 2}}
    d_j = \frac{{z^i_t}^T z^j_t}{ \| z^i_t\|_{\small 2} \| z^j_t\|_{\small 2}},~j=\{1,...,b\} \setminus i
\end{equation}

Similarly, with cosine similarity $d_j, j=\{1,..., b\} \setminus i$ we build the relative distribution $\bm{q}$:
\begin{equation} \label{eq14}
    q_j = \frac{exp(d_j/\tau)}{\sum_{l\in \{1,..., b\} \setminus i} exp(d_l/\tau)}, j=\{1,..., b\} \setminus i
\end{equation}
To quantify the uncertainty of relative distribution, i.e., how confident when we extract the neighbors, we define a similar confidence measure $c$ for in-batch samples:
\begin{equation} \label{eq15}
    c = 1 - \frac{H(\bm{q})}{log(b - 1)}
\end{equation}
and the corresponding adaptive soft label as:
\begin{equation} \label{eq16}
    y_j =\left\{
\begin{aligned}
& 1, ~j = i \\
& max(1, c{\cdot}K{\cdot}p_k), ~j\neq i 
\end{aligned}
\right.
\end{equation}
\end{document}